\documentclass[11pt,a4paper]{article}
\usepackage[hyperref]{acl2021}
\usepackage{times}
\usepackage{latexsym}

\usepackage{booktabs}
\usepackage{microtype}

\aclfinalcopy 

\usepackage{graphicx}
\usepackage{rotating}
\usepackage{tikz}
\usepackage{highlight}

\usepackage{multirow}
\usepackage{amsmath}
\usepackage{graphics}
\usepackage{tabularx}
\newcolumntype{s}{>{\hsize=.24\hsize}X}
\newcolumntype{t}{>{\hsize=.32\hsize}X}
\usepackage{tabu}
\usepackage{booktabs}
\usepackage{highlight}

\usepackage{graphicx}
\usepackage{adjustbox}
\usepackage{tikz}
\usepackage{xspace}
\usepackage{caption}

\usepackage{color, colortbl}
\definecolor{Gray}{gray}{0.95}
\definecolor{DarkGray}{gray}{0.8}

\usepackage{pgfplots}
\pgfplotsset{compat=1.14}
\usepackage{subcaption}
\usepackage{bm}
\renewcommand{\vec}[1]{\bm{#1}}

\title{Contextualizing Variation in Text Style Transfer Datasets}

\author{
Stephanie Schoch$^{ }$\quad Wanyu Du$^{ }$\quad Yangfeng Ji$^{ }$\\
$^{ }$ Department of Computer Science\\University of Virginia\\ Charlottesville, VA 22904\\
\texttt{\{sns2gr,wd5jq,yangfeng\}@virginia.edu}\\
}

\date{}

\begin{document}
\maketitle
\begin{abstract}
Text style transfer involves rewriting the content of a source sentence in a target style. Despite there being a number of style tasks with available data, there has been limited systematic discussion of how text style datasets relate to each other. This understanding, however, is likely to have implications for selecting multiple data sources for model training. While it is prudent to consider inherent stylistic properties when determining these relationships, we also must consider how a style is realized in a particular dataset. In this paper, we conduct several empirical analyses of existing text style datasets. Based on our results, we propose a categorization of stylistic and dataset properties to consider when utilizing or comparing text style datasets.
\end{abstract}

\section{Introduction}
\label{sec:intro}
The general task of text style transfer involves rewriting source content in a target style. Currently, there are a number of text style transfer tasks with available data, such as formality \citep{Tetreault2018}, bias \citep{pryzant2020automatically}, sentiment \citep{10.1145/2872427.2883037}, humor or romance \citep{gan2017stylenet}, offensiveness, \citep{nogueira-dos-santos-etal-2018-fighting}, authorship or time period \citep{xu-etal-2012-paraphrasing}, and personal attributes \citep{kang-etal-2019-male}.
While these specific tasks are often modeled in isolation, the general task definition remains consistent. As such, a natural question arises of what the relationship is between the stylistic variation of specific tasks. 

Stylistic variation can arise from a number of factors such as communicative intent, topic, and speaker-receiver dynamics \citep{biber2019register}, yet within the task of text style transfer, our view of a style is constrained to the context of each specific dataset. Therefore, understanding the tasks as well as the relationships between different tasks requires considering the stylistic properties and potential contextual and social factors \citep{hovy-yang-2021-importance, hovy-2018-social} underpinning them, as well as the dataset characteristics \citep{bender2018data} and intersection of influences giving rise to the realization of style within a dataset.

From an application standpoint, considering these influences can provide a more comprehensive understanding of important textual features. There is a body of work already looking at how to identify generic features to increase target task performance \citep{li-etal-2019-domain} or to compute similarity of textual features to select data for transfer learning \citep{ruder-plank-2017-learning}. In the context of text style transfer, these approaches first require understanding what features should be shared across tasks. For example, \citet{zhang-etal-2020-parallel} leveraged the stylistic features shared between grammatical error correction data and formality to increase model performance on formality transfer datasets.

\begin{table*}[th!]
\resizebox{\textwidth}{!}{%
\centering
\small
\begin{tabular}{lllllll}
  \toprule
  \multirow{2}{*}{\textbf{Dataset}} & \multirow{2}{*}{\textbf{Stylistic Task}} & \multirow{2}{*}{\textbf{Domain}} & \multirow{2}{*}{\textbf{Annotation}} & \multicolumn{3}{c}{\textbf{Size}}\\ 
  \cmidrule{5-7}
  &&&& Train & Dev & Test\\
  \midrule
  Flickr & Romantic$\rightarrow$Humorous & Image Captions & Manual & 6k & 500 & 500\\
  \midrule
  Shakespeare & Shakespeare$\rightarrow$Modern & Literature, SparkNotes & Automatic & 18.4k & 1.2k & 1.5k\\
  \midrule
  GYAFC-FR & Informal$\rightarrow$Formal & Yahoo Answers (Online) & Manual & 52k & 2.8k & 1.3k\\
  \midrule
  GYAFC-EM & Informal$\rightarrow$Formal & Yahoo Answers (Online) & Manual & 52.6k & 2.9k & 1.4k\\
  \midrule
  Biased-word & Subjective$\rightarrow$Neutral & Wikipedia (Online) & Automatic & 53.8k & 700 & 1k\\
  \midrule
  Fluency & Disfluent$\rightarrow$Fluent & Telephone Conversations & Manual & 173.7k & 10.1k & 7.9k\\
  \bottomrule
\end{tabular}}
\caption{\label{dataset-table} An overview of the datasets used for exploratory analyses. Task describes the source-target direction used in our experiments and domain and annotation show general categorizations. Size provides statistics of the data splits, with standard, pre-existing data splits used when available.}
\end{table*}

In addition to textual features such as stylistic properties, existing work also suggests that context of dataset creation should be taken into account when identifying compatible data or assessing possible out-of-distribution generalizability. For example, the similarity between how sentiment information is reflected in different domains affects adaptation performance \citep{li-etal-2019-domain}, and many models can achieve high performance on natural language inference tasks through task-limiting annotation artifacts \citep{gururangan-etal-2018-annotation, poliak-etal-2018-hypothesis}. In other words, factors such as data source and annotation method can create underlying textual features that can impact performance and limit generalizability. Thus, in combination, these existing works on leveraging inherent stylistic similarities \citep{zhang-etal-2020-parallel} or similar style-representations in different dataset domains \citep{li-etal-2019-domain}, as well as identifying task-limiting dataset properties \citep{gururangan-etal-2018-annotation, poliak-etal-2018-hypothesis} indicate that analysis of both stylistic properties and dataset characteristics, as well as the potential interdependencies between them, is warranted. 

In this paper, we consider two primary categories of textual variation within the context of text style transfer: \textbf{stylistic characteristics} and \textbf{dataset characteristics}. We perform a series of empirical analyses to demonstrate the visible influence of both style and dataset characteristics on the performance of text style transfer models. Then, we present a categorization of style and dataset properties for consideration when utilizing or comparing style transfer datasets. 
Finally, we discuss the downstream applications for contextualizing variation in text style datasets, including multi-task learning, data selection, and generalizability. Our work and suggestions fall within the context of and align with recent work on incorporating social factors in natural language processing systems \citep{hovy-yang-2021-importance} and characterizing datasets \citep{bender2018data}.

\section{Empirical
Analyses}\label{sec:exploratoryexperiments}
As an exploratory step, we question whether we can distinguish differences arising from style or dataset properties when comparing empirical results across datasets. We identify a set of aligned English datasets used for supervised text style transfer that exhibit differences ranging from style, annotation method, and domain. We further restrict our selection to datasets in which a single stylistic attribute is transferred between classes. Specifically, we look at \textbf{GYAFC-EM} \& \textbf{GYAFC-FR} \citep{Tetreault2018}, \textbf{Shakespeare} \citep{xu-etal-2012-paraphrasing}, \textbf{Biased-word (Bias)} \citep{pryzant2020automatically}, \textbf{Fluency} \citep{wang2020multi, godfrey1992switchboard}, and \textbf{Flickr} \citep{gan2017stylenet}. We provide dataset overviews in Table \ref{dataset-table}, with detailed dataset descriptions provided in Appendix \ref{sec:datasets}. We perform a preliminary qualitative analysis to get an initial impression of the data differences.

\paragraph{First Impression of Data: } Of the six datasets, four were manually annotated and two were automatically annotated. For manually annotated datasets, GYAFC-EM and GYAFC-FR utilized crowdsourced rewrites, Flickr utilized crowdsourced sentences with only visual context shared between annotators, and Fluency utilized expert annotations of the target attribute. Both automatically annotated datasets (Bias, Shakespeare) were created through identification of existing data sources. While each style task is unique (other than two domains of GYAFC for formality), in terms of style we observe that Shakespeare has a significantly different temporal context than all other datasets, and Fluency involves a stylistic attribute that, ideally, the sentence pairs in all other datasets should possess.\footnote{Fluency is frequently a criteria used in text style transfer evaluation \citep{mir-etal-2019-evaluating, briakou2021review, prabhumoye-etal-2018-style}.}

Beyond our qualitative observations, we perform an exploratory multi-task learning experiment, described in the following subsection.

\subsection{Multi-Task Learning}
As a toy experiment, we ask the question ``\textit{What would our results look like if we naively train on all style transfer tasks, with no considerations beyond the fact that the tasks share a general task definition?}\footnote{The general task definition is rewriting the source content of a text in a target style (see section \ref{sec:intro})} We essentially ignore all considerations for style or dataset properties. Our expectation is that negative transfer will occur due to the lack of consideration for factors such as domain \citep{pan2009survey, li-etal-2019-domain}\footnote{Negative transfer occurs when transferred knowledge negatively impacts target performance \citep{pan2009survey}.}, but we are interested in whether all tasks share similar performance patterns or if performance on any tasks diverge from the overall set. If the latter, is there any intuitive explanation for the divergences?

We further expect that the degree of negative transfer will be impacted by the degree of difference of stylistic or data properties, relative to the full set of pre-training datasets. Specifically, we anticipate some level of alignment with our initial impression of the data: the alternate temporal context of Shakespeare may increase degree of negative transfer, yet the inherent stylistic connection with Fluency may lessen the degree of negative transfer. 

\paragraph{Experimental Setup} We utilize two experimental settings: GPT-2 directly fine-tuned on each dataset, and GPT-2 with multi-task pre-training on all datasets followed by fine-tuning on each target dataset. For both settings, we initialize GPT-2 with the pre-trained parameters from \citet{radford2019language}. For our multi-task experimental setup, we follow prior works \citep{liu-etal-2015-representation, liu-etal-2019-multi-task, JMLR:v21:20-074} to perform multi-task learning for the baseline GPT-2 model \citep{wang-etal-2019-harnessing}: we initialize GPT-2 with the pre-trained parameters from \citet{radford2019language}, then we jointly pre-train on all style tasks in a supervised manner and fine-tune on each individual style transfer task. \footnote{GPT-2 models were each trained on a single NVIDIA GTX 1080 Ti GPU.}

For multi-task learning, we construct our pre-training dataset by randomly shuffling the training examples from all datasets. During pre-training, each training example from each individual task is seen at least once per epoch. All of the training examples in the largest dataset are seen exactly once per epoch, while all training examples for the smallest dataset are seen multiple times per epoch (proportional to the ratio between the training set size of the largest-scale task and the smallest-scale task). For the fine-tuning step, we leverage the multi-task pre-trained model and further fine-tune on each individual supervised task, saving the model with the lowest validation set loss as our final model for evaluation.

\begin{table}[t]
\resizebox{\columnwidth}{!}{%
\centering
\small
\begin{tabular}{lllll}
\toprule
\textbf{Dataset} & \textbf{Task} & \textbf{BLEU-og} & \textbf{BLEU-mt} & \textbf{\%og}\\
\midrule
Shakespeare & shake2mod & 24.47 & 11.33 & 0.463\\
Fluency & dis2fl & 96.59 & 96.69 & 1.001\\
Flickr & rom2fun & 8.14 & 7.18 & 0.882\\
GYAFC-EM & inf2fr & 69.96 & 65.16 & 0.931\\
GYAFC-FR & inf2fr & 75.16 & 74.72 & 0.994\\
Biased & subj2neut & 93.73 & 93.41 & 0.996\\
\bottomrule
\end{tabular}}
\caption{\label{mt-table} Experiments conducted using GPT-2, where BLEU-og represents directly fine-tuning the original GPT-2 on the target task, BLEU-mt represents multi-task pre-training using all datasets and fine-tuning on the target task, and \%og represents the relative performance of multi-task pre-training in comparison to the performance of the original GPT-2 (computed by dividing BLEU-mt by BLEU-og).
}\end{table}

\paragraph{Results} We report BLEU \citep{papineni2002bleu} in Table \ref{mt-table} as a measure of content preservation.\footnote{We use the BLEU implementation from \citet{koehn-etal-2007-moses}.} We compare the performance between directly finetuning the original GPT-2 on the target task (BLEU-og) and firstly multi-task pretraining the original GPT-2 then fine-tuning it on the target task (BLEU-mt). 

Negative transfer is identified as a performance drop in BLEU-mt, i.e. \%og $<$ 1.00. Since the style transfer datasets in use are diverse across domain and stylistic properties, we expect negative transfer to occur in the multi-task learning setting. However, we are specifically looking at the overall performance pattern as an initial step in determining what properties may underlie such differences, which should be accounted for in a taxonomy.

While most tasks perform within a 12\% margin below the original GPT-2 performance, we observe two divergences: with multi-task learning, the Shakespeare-to-modern task performed at less than 50\% of the original GPT-2 performance, and the disfluent-to-fluent task experienced a slight performance increase. Performance on Fluency exceeded our initial expectation that the degree of negative transfer would simply be lower compared to other datasets, but overall the divergences with Shakespeare and Fluency match our expectations based on our initial impression of the data style differences. Specifically, we attribute the performance drop on the Shakespeare dataset to limited suitability for combining the data sources likely due to the stylistic attribute pertaining to different temporal context, and we attribute the Fluency dataset performance increase to high suitability for combining the data sources likely due to its stylistic attribute pertaining to a textual criteria that is assumed to be inherent to the other data.

With regard to dataset differences, we note the potential impact of dataset size on performance: to maintain consistency of the model architecture, we utilize the same model configuration with GPT-2 across datasets and experimental settings. In the case of performance on the Flickr dataset (see Table \ref{dataset-table}), it is possible that such a model configuration may overfit on the dataset. However, this alone fails to account for our observations of performance pattern divergences.

Beyond overall pattern, we observe an unexpectedly wide range of BLEU scores across datasets, which we expect could be attributable to differences in either dataset creation or style.  There may be stylistic differences in how style information is encoded that impact content preservation. For example, some styles may have more words that encode both style and content information which may increase the difficulty of content retention \citep{cao-etal-2020-expertise}, yet other styles may be characterized by stylistic attributes encoded in only a few key words or phrases \citep{fu-etal-2019-rethinking}. However, these differences may also be attributable to dataset creation. We expect that if the attribute-encoding words are constrained to a few words or phrases as a property of a style itself, then a dataset's style classes should be highly distinguishable using lexical features; in other words, the decision boundary when classifying styles should stay at the lexical level \citep{fu-etal-2019-rethinking}. 

To test these hypotheses and help explain the range of BLEU scores, we perform two complementary experiments. First, we compute sentence similarity metrics averaged over each dataset to 1) identify if there is a relationship between BLEU scores and baseline sentence pair similarities, and 2) identify datasets with high similarity across class boundaries that constrain stylistic attributes to a few words or phrases. Second, we perform classification and ablation studies using a set of linguistic features defined on each dataset. For datasets with high sentence similarities, if a style can be well-represented by a few style-encoding words or phrases, then we expect high classification performance using only lexical features. Conversely, if a style cannot be isolated to a few words and phrases, we expect low classification performance using lexical features alone, in which case a high sentence similarity is likely attributable to dataset properties rather than inherent style properties.

\begin{table}[t]
\resizebox{\columnwidth}{!}{%
\centering
\small
\begin{tabular}{lllll}
\toprule
\textbf{Dataset} & \textbf{JS $\uparrow$} & \textbf{LD $\downarrow$} & \textbf{LD-norm $\downarrow$}& \textbf{F1-Score $\uparrow$}\\
\midrule
Shakespeare & 0.0845 & 14.79 & 0.9029 & 0.0583\\
Fluency & 0.9941 & 0.366 & 0.0271 & 0.9751\\
Flickr & 0.2257 & 11.92 & 0.7728 & 0.3623\\
GYAFC-EM & 0.4471 & 7.924 & 0.5616 & 0.4207\\
GYAFC-FR & 0.4565 & 7.723 & 0.5375 & 0.4500\\
Biased & 0.9137 & 2.529 & 0.0763 & 0.9689\\
\bottomrule
\end{tabular}}
\caption{\label{sent-sims-table}Jaccard Similarity (JS), Levenshtein Distance (LD), normalized Levenshtein Distance (LD-norm), and F1-Score. Sentence similarity measures quantify the distance between target and source for the training sets with arrows indicating direction for more similar sentences.} 
\end{table}

\subsection{Similarity Metrics} We calculate token-based Jaccard Similarity, token-based Levenshtein distance, and $F_1$-score between the source and target training sets.  We also report Levenshtein distance normalized by sentence length,  $LD_{norm}(\vec{s},\vec{t})=\left(\frac{LD(\vec{s},\vec{t})}{\max{|\vec{s}|,|\vec{t}|}}\right)$ where $LD(\vec{s},\vec{t})$ is the Levenshtein distance, $\vec{s}, \vec{t}$ refer to sentences in a sentence pair, and $|\cdot|$ refers to the number of tokens in a sentence. Scores are reported in Table \ref{sent-sims-table}.\footnote{We do not distinguish between source and target direction as these metrics are symmetric in our setting (see Appendix \ref{sec:similarity-app}).}

We see some relationships between similarities in Table \ref{sent-sims-table} and GPT-2 performances in Table \ref{mt-table} in that the datasets with the lowest BLEU scores (Shakespeare and Flickr) have the lowest baseline similarities, and the datasets with the highest BLEU scores (Fluency and Bias) have the highest baseline similarities. We therefore can identify the Fluency and Bias datasets as being of particular relevance for the lingusitic features analysis. Specifically, our hypothesis is that if the Bias and Fluency styles can truly be isolated to few words as the sentence similarities would suggest, then the classification performance should be high using only lexical features. In contrast, if the dataset properties influence variation through constrained stylistic representation, then we expect low classification accuracy using lexical features.

\subsection{Linguistic Features Analysis}
We define linguistic features to refer to properties characterizing textual variation primarily at the lexical or syntactic level, where the ``other'' category in Table \ref{features-table} indicates features that may capture slight semantic variation (subjectivity) or reflect overall lexical tendencies (bag-of-words). Features are adopted from prior works \citep{pavlick-tetreault-2016-empirical, abu-jbara-etal-2011-towards, roemmele2017evaluating} and listed in Table \ref{features-table}, with further description in Appendix \ref{ling-feat}.

\begin{table}[t]
\resizebox{\columnwidth}{!}{%
\centering
\fontsize{16}{18}\selectfont 
\begin{tabularx}{\textwidth}{lX}
\hline
\rowcolor{DarkGray}
\textbf{Group} & \textbf{Features} \\
\hline
Lexical Complexity & Average word length, average syllable count (with \& without stopwords) \\
Readability & \# complex words ($\geq3$ syllables)*, Flesch Reading Ease Score, Flesch-Kincaid Grade Level \\
Lexical Diversity & Unique unigrams \& bigrams, with punctuation removed*\\
\hline
\rowcolor{Gray}
POS tags & Universal POS tag distribution*, Penn Treebank POS tag distribution* \\
\rowcolor{Gray}
Sentence length & Sentence length (words \& total tokens) \\
\rowcolor{Gray}
Phrases & \# noun phrases*, \# verb phrases*, average length of noun phrases*, average length of verb phrases*, \# dependent clauses*, average length of dependent clauses* \\
\hline
Subjectivity & \# 1st, 2nd, \& 3rd person pronouns*, Subjectivity \& Sentiment polarity according to TextBlob sentiment module \\
Bag-of-Words & Bag-of-words feature representation\\
\hline

\end{tabularx}}
\caption{\label{features-table} Linguistic feature groups: lexical (top), syntactic (gray in middle), and other (bottom). Features features denoted with an asterisk (*) are normalized by sentence length.}
\end{table}

We train logistic regression classifiers with $\ell1$-regularization and feature scaling on the full feature set for each text style dataset. Next, we train and subsequently test classifiers with all features ablated except the specified subset, and identify important features as those with minimal relative performance drop compared to full-feature classification accuracy. Results are shown in Table \ref{heatmap}. We further quantify the magnitude of variation by computing the Jensen-Shannon (JS) divergence for each feature, and indicate the cells corresponding to features with divergences $\geq 0.075$ in Table \ref{heatmap} in bold.\footnote{Table \ref{divergence-table} in Appendix \ref{div-app} shows a JS-divergence heatmap.}

\begin{table}[t]
\resizebox{\columnwidth}{!}{%
    \centering
    \begin{tabular}{r*{8}{r}}
        \multicolumn{1}{c}{} &
        \multicolumn{1}{c}{Flick} &
        \multicolumn{1}{c}{Shake} &
        \multicolumn{1}{c}{GY-FR} &
        \multicolumn{1}{c}{GY-EM} & \multicolumn{1}{c}{Bias} & \multicolumn{1}{c}{Flu.}\\ 
        FF&\gradient{75.6}&\gradient{76.1}&\gradient{80.7}&\gradient{80.9}&\gradient{63.5}&\gradient{55.3}\\ 
        LexC&\textbf{\gradient{51.7}}&\gradient{62.2}&\textbf{\gradient{65.6}}&\gradient{64.4}&\gradient{52.6}&\gradient{50.7}\\ 
        Read&\textbf{\gradient{55.7}}&\gradient{52.1}&\textbf{\gradient{62.1}}&\gradient{63.3}&\gradient{52.0}&\gradient{51.0}\\ 
        LexD&\gradient{52.4}&\gradient{49.6}&\gradient{51.2}&\gradient{52.0}&\textbf{\gradient{50.4}}&\textbf{\gradient{54.4}}\\ 
        UPOS&\textbf{\gradient{59.4}}&\gradient{59.3}&\gradient{62.3}&\textbf{\gradient{60.8}}&\gradient{54.4}&\gradient{51.6}\\ 
        XPOS&\gradient{62.3}&\gradient{59.7}&\gradient{65.1}&\gradient{66.1}&\gradient{55.0}&\gradient{51.7}\\
        SenL&\textbf{\gradient{51.8}}&\textbf{\gradient{56.7}}&\gradient{56.2}&\gradient{51.7}&\gradient{50.3}&\gradient{51.0}\\
        Phr&\textbf{\gradient{54.2}}&\gradient{58.2}&\gradient{53.6}&\gradient{53.4}&\gradient{52.9}&\gradient{51.8}\\
        Sub&\textbf{\gradient{60.5}}&\textbf{\gradient{60.4}}&\gradient{51.7}&\gradient{52.9}&\gradient{57.0}&\gradient{50.4}\\
        BoW&\gradient{74.2}&\gradient{72.4}&\gradient{71.5}&\gradient{71.7}&\gradient{62.2}&\gradient{50.3}\\ 
    \end{tabular}}

\caption{\label{heatmap}Classification accuracy using linguistic feature groupings described in Table \ref{features-table}, with Full Features (FF) indicating the entire suite of features. Classification accuracy for features with Jensen-Shannon divergences $\geq 0.075$ are in bold.}
\end{table}

Datasets with the lowest BLEU scores (Flickr and Shakespeare) have more distributed salient class features across linguistic levels, further reflected in a higher number of features with large divergence magnitudes ($\geq 0.075$). For the high BLEU and sentence similarity datasets of interest (Bias, Fluency), the inverse of this is true. For Bias and Fluency we see consistently low classification performance across ablations, including the lexical feature ablations. These results support our hypotheses and further suggest that neither stylistic differences nor dataset characteristics alone can be used to relate text style datasets. Rather, both influences as well as their interactions require consideration.

In the following section, we propose a taxonomy of style and dataset property categories that can contribute to variation in text style transfer datasets. Additionally, we note that when introducing these properties, we view style \textit{as the targeted stylistic property within the context of a text style dataset}.

\section{Variation From Style and Data Properties}\label{sec:framework}
\begin{figure*}
    \centering
    \includegraphics[width=\textwidth, angle=0]{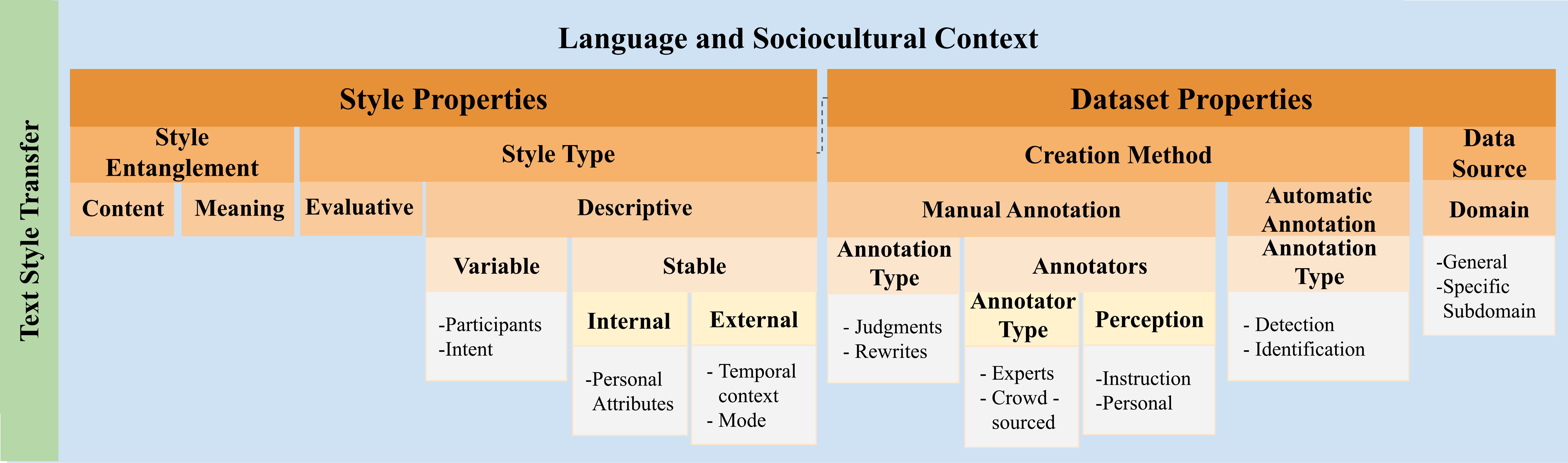}
    \caption{Framework overview visualizing style and dataset properties discussed throughout section \ref{sec:framework}. Grey boxes indicate example considerations within each category. We contextualize both style and dataset properties within language and sociocultural context as all language is implicitly reflective of these influences \citep{hovy-yang-2021-importance}.}
    \label{fig:stg_framework}
\end{figure*}

Our empirical analyses demonstrate the visible influence of both style and dataset properties on how a style is represented in a given dataset. In addition to brief mentions of influences of dataset creation in section \ref{sec:intro}, we can identify an intuitive reason for these dual influences. While linguistic approaches exist to analyze textual variation \citep{halliday2013halliday, holmes2017introduction, biber2012register}, we suggest that the processes of linguistic-based stylistic analysis and text style transfer typically occur in inverse directions: 
linguistic analysis may work from human-written text and then analyze stylistic variation, whereas text style transfer may work from pre-existing ideas of targeted stylistic variation and then create datasets of human-written text that meet stylistic expectations. In other words, to create a text style transfer dataset or train a text style transfer model, the researcher should have a notion of the desired style against which to judge the resulting artifact. Intuitively, this process can lead to process-attributable variation secondary to and alongside the intended stylistic variation. 

Based on our results and observations, we consider \textbf{\textit{stylistic properties}} as properties influencing textual variation that are inherent to a particular style and \textbf{\textit{dataset properties}} as factors influencing textual variation due to how a particular dataset was created. We detail style and dataset properties in the following subsections and visualize the major distinctions in Figure \ref{fig:stg_framework}.

\subsection{Stylistic Properties}
We group stylistic properties under two broad categories: \textit{style entanglement} and \textit{style type}.

\subsubsection{Style Entanglement}
Although some recent approaches to style transfer model style and content words separately \citep{li-etal-2018-delete}, or try to disentangle style and content representations \citep{john-etal-2019-disentangled, kazemi2019style}, this approach may be less effective when used to transfer styles in which a higher ratio of words embed both style and content information. We can consider this ratio of dual-embedding a property inherent to the style. Specifically, we can consider how entangled the style and the content or semantic meaning is, where \textit{content entanglement} refers to whether changes to the style result in additions or reductions in the total content details, and \textit{meaning entanglement} refers to whether changes to the style can retain the content details but alter the semantic meaning. As an example of this distinction, sentiment transfer, which has been regarded previously as transfer between negative and positive style \citep{shen2017style, prabhumoye-etal-2018-style} alters semantic meaning while retaining most content, yet transferring between styles such as expert-to-layman can retain meaning but lead to content detail reductions due to the difficulty of preserving content from professional sentences \citep{cao-etal-2020-expertise}. 

\subsubsection{Style Type}
Style can refer to the \emph{individuating} sense or \emph{evaluative} sense of a text \citep{crystal_investigating_1969}. We refer to \textbf{\textit{evaluative styles}} as styles distinguished by general properties that address overall textual quality corresponding with rules of usage and composition, effectiveness of expression \citep{strunk_elements_1999} or based on overall quality evaluation and judgments \citep{williams_style_2017}. Stylistic variation occurs solely along evaluative lines, independent of situational context or language choice. From our empirical experiments, we can consider the Fluency dataset representative of a dataset in which the transferred stylistic attribute refers to an evaluative sense of style.

We consider \textbf{\textit{descriptive styles}} as distinguished by stylistic properties that characterize textual variation through influences such as the underlying communicative intent, the situational or social factors influencing language choice, and the attributes of the producer of the text. We can further differentiate descriptive styles by the stability or variability of the targeted stylistic property.

\paragraph{Stability of Targeted Style Properties}
On one end of the spectrum variable stylistic properties (high variance, low stability) are characterized by dynamically shifting language to convey information a certain way, which may be reflective of factors such as the underlying intent in producing the text or the social dynamics of a situation. For example, politeness can shift based on social dynamics such as social distance and relative power between participants \citep{brown1987politeness} independently of the directness of communication, such as formality \footnote{Formality is closely related to politeness \citep{kang2019xslue}} in email \citep{peterson-etal-2011-email}. From our empirical experiments, we consider Flickr, GYAFC, and Bias as reflective of variable targeted properties.

At the other end of the spectrum, more stable targeted stylistic properties (low variance, high stability) remain more consistent across social situations and arise from relatively stable internal or external context. These may reflect internal context such as the personal attributes of the producer of text \citep{kang-etal-2019-male}, or external context such as the temporal context at time of text production or stylistic properties inherent to the mode of distribution. Example datasets include the \textsc{pastel} dataset \citep{kang-etal-2019-male} annotated for personal attributes such as gender and age group, and the Shakespeare dataset \cite{xu-etal-2012-paraphrasing} which can be considered reflective of authorship \citep{xu-2017-shakespeare} or temporal context.\footnote{Regarding distribution mode, \citet{abu-jbara-etal-2011-towards} suggested a set of linguistic features differentiating written and audio styles.}

\subsection{Dataset Properties}
While in the previous section we discussed properties inherent to specific styles, in this section we discuss properties of datasets to which textual variation is attributable. We identify the broad categories of properties due to \textit{creation method} and \textit{data source}. In this context, creation method refers to the general method of creating sentence pairs (automatic or manual annotation, as well as any properties arising from utilizing a specific method, such as influences of annotator background or perceptions) and data source refers to characteristics (such as domain) from where the source data was collected. We provide more detailed discussion in the following subsections.

\subsubsection{Creation Method}
Generally speaking, datasets can be created via manual annotation, such as through judgments or rewrites, or via automatic annotation, such as through filtering data that has a target attribute (i.e., detection with a classifier). With particular attention on manual annotation, in addition to potential generalizability-limiting data properties arising from artifacts of the \textit{annotation method} and \textit{annotation type} (\citep{geva-etal-2019-modeling}, also, see section \ref{sec:intro}), the \textit{annotators} themselves can influence stylistic variation. For example, model performance has been improved by incorporating annotator identifiers as features \citep{geva-etal-2019-modeling} and by augmenting machine translation models with distinct translator styles identifiable in the training data \citep{wangtowards}. In the case of \citet{wangtowards}, using annotator styles resulted in BLEU score variations of up to $+4.5$ points.

Underlying these influences, annotator properties that may give rise to textual variation could include the background of the annotator such as experts or crowd-sourced workers, and the perception the annotators have of the style task. Similar to human evaluation of outputs, perception may arise due to personal understanding or the wording of instructions presented.\footnote{\citet{schoch-etal-2020-problem} discuss potential influences of framing effects of questions or instructions on results in human evaluation of outputs, and we suggest similar effects could influence dataset properties resulting from annotation of inputs.}.

\paragraph{Data Source - Domain:}
Differences in domain can be reflected in entirely different word meanings and contexts of use \citep{li-etal-2019-domain}, as well as different manners of encoding attribute information such as sentiment \citep{blitzer-etal-2007-biographies, li-etal-2019-domain}. In addition to differences of a single style between domains, the domains themselves have different levels of stylistic diversity \citep{kang2019xslue}. Further, while the properties characterizing a style may be inherent to how a style is realized \textit{within} a domain, there is a distinction in how the style is reflected \textit{between} domains that necessitates domain being considered as a dataset property influencing variation in text style datasets.

\section{Interplay Between Style and Data Properties}\label{sec:dependence}
\citet{bender2018data} proposed data statements for documenting dataset contextual factors such as language variety, speaker demographics, annotator demographics, speech situation, and text characteristics (e.g. genre, topic). The style and dataset properties we discuss as potentially contributing to variation in text style transfer datasets show some alignment with those proposed for data statements as such factors contribute to linguistic variation in a general sense. However, our categorization specifically operates within the context of text style transfer datasets for which there are unique considerations and important distinctions between sources of variation and downstream implications or applications. 

In the previous subsections, we discussed style properties and dataset properties to which variation in text style transfer datasets can be attributed. In this section, we discuss the interdependence of style and data properties in text style transfer datasets in terms of  context-dependence of and interactions between sources of variation.

\paragraph{Style and Data Property Interactions}
While we previously considered the potential impact of both style and dataset characteristics independently, these characteristics may have underlying interactions and influences on one another. Specifically, certain types of stylistic properties may be more or less amenable to certain dataset creation methods or sources, and vice versa.

With regard to the stability of stylistic properties, dataset properties such as annotation method may be indirectly influenced when transferring across relatively stable stylistic properties. For example, machine translation models have been found to exhibit stylistic bias through reflecting demographically-biased training data \citep{hovy-etal-2020-sound}. While this demonstrates that the demographics of annotators can serve as an important dataset characteristic, it also demonstrates the potential to transfer across relatively stable stylistic properties, such as personal attributes \citep{kang-etal-2019-male}. However, as the stylistic properties are inherent to the annotator, there may be constraints on dataset creation through manual data annotation, such as potential limitations and additional considerations for using methods such as human judgments. This underscores additional considerations for and potential challenges of selecting data from two styles that may have underlying influences on how datasets are constructed.

\paragraph{Context-Dependence of Variation}
Relatedly, contextual considerations come into play with respect to the the Shakespeare to Modern English style transfer task, a dataset also reflective of transfer across stable, contextual boundaries. The Shakespeare to Modern English transfer task can be considered as transferring across temporal context, or as the characteristic style of a single author \citep{xu-2017-shakespeare}. In this case, while an influence of sociocultural context is apparent when considering the original data sources, the targeted stylistic variation occurs across such context boundaries. Thus, source of variation for textual features arising from external context lies with whether the intent is present for a dataset to represent a transfer across context boundaries, rather than an artifact reflecting specifics of dataset creation. This is illustrated in Figure \ref{fig:stg_framework} as a dashed line connecting style type to dataset properties. 

With further regard to dataset creation, it is important to acknowledge that while we consider many properties arising from social influences as dynamic and variable influences giving rise to particular \textit{styles}, a dataset will indirectly and inadvertently reflect such social context during creation to some degree. As such, we also must consider social factors \textbf{not} related to the actual targeted style, but rather arising from the dataset creation process. As an example of this consideration, we can't simply say two sentiment datasets from the same general domain (such as restaurant reviews) are equivalent if one was constructed with reviewers who had anonymity (in a sense mitigating some of the direct social pressure or influence) and the other was constructed with reviewers who were not anonymous and were thus subject to increased social pressure. By understanding both data and style differences and their interactions within a particular context, these potential differences or hidden influences can be more easily identified. In summary, the interactions between style and data properties are complex. While we have suggested interactions between context and sources of influence, there are likely correlations that exist based on sources of variation which future work can investigate.

\section{Influences and Applications}\label{sec:framework-usefulness}
In the previous sections, we demonstrated visible influences of style and dataset properties on performance, categorized a set of style and dataset properties for consideration, and discussed the potential interactions between sources of variation. We conclude by discussing several applications of understanding the sources of variation in text style transfer datasets. Specifically, we look at multi-task learning, domain adaptation, and generalizability.

\paragraph{Multi-Task Learning and Domain Adaptation}
Multi-task learning aims to jointly train a model with auxiliary tasks to complement learning of the target task. When determining which auxiliary objectives to incorporate, multi-task learning for various NLP tasks has been shown to benefit from knowledge about both \emph{dataset characteristics} and \emph{stylistic properties}. For example, multi-task learning performance gains for NLP tasks such as POS tagging and text classification are predictable from dataset characteristics \citep{kerinec-etal-2018-deep, bingel-sogaard-2017-identifying}. With regard to stylistic properties, within the context of multi-task learning for style transfer \citet{zhang-etal-2020-parallel} achieved performance gains by leveraging an intuitive stylistic connection between formality data and grammatical error correction data.\footnote{Other styles, such as impoliteness and offense, are also highly dependent on each other \citep{kang2019xslue}}.

While multi-task learning can be viewed as a form of parallel transfer learning, we can view domain adaptation as a form of sequential transfer learning and look at similar applications of contextualizing stylistic variation. \citet{li-etal-2019-domain} found that leveraging generic style and content information outperformed generic content information alone for domain adaptation, however, the closeness of sentiment information (target attribute) in the source and target domains impacted performance. In other words how the style was reflected in the particular dataset (i.e., a dataset characteristic) was related to the benefit provided by the adaptation. Based on the combined evidence in this section, we can thus support applying analysis of both style and dataset properties for transfer learning data selection, including multi-task learning and domain adaptation, in text style transfer. We suggest that the taxonomy presented in this paper can assist exploration of systematic data selection methods in these and related application areas.

\paragraph{Generalizability}
One of the underlying motivations for pursuing multi-task learning and domain adaptation is the issue of generalizability. In the context of style transfer, we can consider generalizing a model for one style across different data distributions with the same stylistic attribute, or across similar domains yet different stylistic attributes. In either case, how the model learns to represent the generic style or content information is vital for successful transfer. As we've demonstrated throughout prior sections, considering both style and dataset properties can aid in identifying sources from which possible issues may arise in terms of along which dimensions stylistic attributes may significantly differ, or which artifacts or influences of dataset creation may influence generalizability secondary to any stylistic considerations. Considerations to this end may prove beneficial both in the dataset creation process as well as when considering how a model may perform beyond a specific dataset.

\section{Conclusion}
In this paper, we conducted a set of exploratory analyses to assess the visibility or influence of both style and dataset characteristics on text style transfer. Based on these observations, we proposed a categorization of stylistic and dataset properties that can contribute to variation in text style transfer datasets and described the applications in which these properties may be influential, limiting, or leveragable.

\section*{Acknowledgements}
We thank the anonymous reviewers for their helpful comments and suggestions. We also thank Diyi Yang and Jingfeng Yang for a series of helpful discussions.

\bibliographystyle{acl_natbib}
\bibliography{anthology,tst}

\begin{thebibliography}{60}
\expandafter\ifx\csname natexlab\endcsname\relax\def\natexlab#1{#1}\fi

\bibitem[{Abu-Jbara et~al.(2011)Abu-Jbara, Rosario, and
  Lyons}]{abu-jbara-etal-2011-towards}
Amjad Abu-Jbara, Barbara Rosario, and Kent Lyons. 2011.
\newblock \href {https://www.aclweb.org/anthology/P11-2043} {Towards style
  transformation from written-style to audio-style}.
\newblock In \emph{Proceedings of the 49th Annual Meeting of the Association
  for Computational Linguistics: Human Language Technologies}, pages 248--253,
  Portland, Oregon, USA. Association for Computational Linguistics.

\bibitem[{Bender and Friedman(2018)}]{bender2018data}
Emily~M Bender and Batya Friedman. 2018.
\newblock Data statements for natural language processing: Toward mitigating
  system bias and enabling better science.
\newblock \emph{Transactions of the Association for Computational Linguistics},
  6:587--604.

\bibitem[{Biber(2012)}]{biber2012register}
Douglas Biber. 2012.
\newblock Register as a predictor of linguistic variation.
\newblock \emph{Corpus linguistics and linguistic theory}, 8(1):9--37.

\bibitem[{Biber and Conrad(2019)}]{biber2019register}
Douglas Biber and Susan Conrad. 2019.
\newblock \emph{Register, genre, and style}.
\newblock Cambridge University Press.

\bibitem[{Bingel and S{\o}gaard(2017)}]{bingel-sogaard-2017-identifying}
Joachim Bingel and Anders S{\o}gaard. 2017.
\newblock \href {https://www.aclweb.org/anthology/E17-2026} {Identifying
  beneficial task relations for multi-task learning in deep neural networks}.
\newblock In \emph{Proceedings of the 15th Conference of the {E}uropean Chapter
  of the Association for Computational Linguistics: Volume 2, Short Papers},
  pages 164--169, Valencia, Spain. Association for Computational Linguistics.

\bibitem[{Blitzer et~al.(2007)Blitzer, Dredze, and
  Pereira}]{blitzer-etal-2007-biographies}
John Blitzer, Mark Dredze, and Fernando Pereira. 2007.
\newblock \href {https://www.aclweb.org/anthology/P07-1056} {Biographies,
  {B}ollywood, boom-boxes and blenders: Domain adaptation for sentiment
  classification}.
\newblock In \emph{Proceedings of the 45th Annual Meeting of the Association of
  Computational Linguistics}, pages 440--447, Prague, Czech Republic.
  Association for Computational Linguistics.

\bibitem[{Briakou et~al.(2021)Briakou, Agrawal, Zhang, Tetreault, and
  Carpuat}]{briakou2021review}
Eleftheria Briakou, Sweta Agrawal, Ke~Zhang, Joel Tetreault, and Marine
  Carpuat. 2021.
\newblock A review of human evaluation for style transfer.
\newblock \emph{arXiv preprint arXiv:2106.04747}.

\bibitem[{Brown et~al.(1987)Brown, Levinson, and
  Levinson}]{brown1987politeness}
Penelope Brown, Stephen~C Levinson, and Stephen~C Levinson. 1987.
\newblock \emph{Politeness: Some universals in language usage}, volume~4.
\newblock Cambridge university press.

\bibitem[{Cao et~al.(2020)Cao, Shui, Pan, Kan, Liu, and
  Chua}]{cao-etal-2020-expertise}
Yixin Cao, Ruihao Shui, Liangming Pan, Min-Yen Kan, Zhiyuan Liu, and Tat-Seng
  Chua. 2020.
\newblock \href {https://doi.org/10.18653/v1/2020.acl-main.100} {Expertise
  style transfer: A new task towards better communication between experts and
  laymen}.
\newblock In \emph{Proceedings of the 58th Annual Meeting of the Association
  for Computational Linguistics}, pages 1061--1071, Online. Association for
  Computational Linguistics.

\bibitem[{Crystal and Davy(1969)}]{crystal_investigating_1969}
David Crystal and Derek Davy. 1969.
\newblock \emph{Investigating {English} {Style}}.
\newblock Indiana University Press, Bloomington \& London.

\bibitem[{Fu et~al.(2019)Fu, Zhou, Chen, and Li}]{fu-etal-2019-rethinking}
Yao Fu, Hao Zhou, Jiaze Chen, and Lei Li. 2019.
\newblock \href {https://doi.org/10.18653/v1/W19-8604} {Rethinking text
  attribute transfer: A lexical analysis}.
\newblock In \emph{Proceedings of the 12th International Conference on Natural
  Language Generation}, pages 24--33, Tokyo, Japan. Association for
  Computational Linguistics.

\bibitem[{Gan et~al.(2017)Gan, Gan, He, Gao, and Deng}]{gan2017stylenet}
Chuang Gan, Zhe Gan, Xiaodong He, Jianfeng Gao, and Li~Deng. 2017.
\newblock Stylenet: Generating attractive visual captions with styles.
\newblock In \emph{Proceedings of the IEEE Conference on Computer Vision and
  Pattern Recognition}, pages 3137--3146.

\bibitem[{Geva et~al.(2019)Geva, Goldberg, and
  Berant}]{geva-etal-2019-modeling}
Mor Geva, Yoav Goldberg, and Jonathan Berant. 2019.
\newblock \href {https://doi.org/10.18653/v1/D19-1107} {Are we modeling the
  task or the annotator? an investigation of annotator bias in natural language
  understanding datasets}.
\newblock In \emph{Proceedings of the 2019 Conference on Empirical Methods in
  Natural Language Processing and the 9th International Joint Conference on
  Natural Language Processing (EMNLP-IJCNLP)}, pages 1161--1166, Hong Kong,
  China. Association for Computational Linguistics.

\bibitem[{Godfrey et~al.(1992)Godfrey, Holliman, and
  McDaniel}]{godfrey1992switchboard}
John~J Godfrey, Edward~C Holliman, and Jane McDaniel. 1992.
\newblock Switchboard: Telephone speech corpus for research and development.
\newblock In \emph{Acoustics, Speech, and Signal Processing, IEEE International
  Conference on}, volume~1, pages 517--520. IEEE Computer Society.

\bibitem[{Gururangan et~al.(2018)Gururangan, Swayamdipta, Levy, Schwartz,
  Bowman, and Smith}]{gururangan-etal-2018-annotation}
Suchin Gururangan, Swabha Swayamdipta, Omer Levy, Roy Schwartz, Samuel Bowman,
  and Noah~A. Smith. 2018.
\newblock \href {https://doi.org/10.18653/v1/N18-2017} {Annotation artifacts in
  natural language inference data}.
\newblock In \emph{Proceedings of the 2018 Conference of the North {A}merican
  Chapter of the Association for Computational Linguistics: Human Language
  Technologies, Volume 2 (Short Papers)}, pages 107--112, New Orleans,
  Louisiana. Association for Computational Linguistics.

\bibitem[{Halliday and Matthiessen(2013)}]{halliday2013halliday}
Michael Alexander~Kirkwood Halliday and Christian~MIM Matthiessen. 2013.
\newblock \emph{Halliday's introduction to functional grammar}.
\newblock Routledge.

\bibitem[{He and McAuley(2016)}]{10.1145/2872427.2883037}
Ruining He and Julian McAuley. 2016.
\newblock \href {https://doi.org/10.1145/2872427.2883037} {Ups and downs:
  Modeling the visual evolution of fashion trends with one-class collaborative
  filtering}.
\newblock In \emph{Proceedings of the 25th International Conference on World
  Wide Web}, WWW '16, page 507–517, Republic and Canton of Geneva, CHE.
  International World Wide Web Conferences Steering Committee.

\bibitem[{Holmes and Wilson(2017)}]{holmes2017introduction}
Janet Holmes and Nick Wilson. 2017.
\newblock \emph{An introduction to sociolinguistics}.
\newblock Routledge.

\bibitem[{Hovy(2018)}]{hovy-2018-social}
Dirk Hovy. 2018.
\newblock \href {https://doi.org/10.18653/v1/W18-1106} {The social and the
  neural network: How to make natural language processing about people again}.
\newblock In \emph{Proceedings of the Second Workshop on Computational Modeling
  of People{'}s Opinions, Personality, and Emotions in Social Media}, pages
  42--49, New Orleans, Louisiana, USA. Association for Computational
  Linguistics.

\bibitem[{Hovy et~al.(2020)Hovy, Bianchi, and
  Fornaciari}]{hovy-etal-2020-sound}
Dirk Hovy, Federico Bianchi, and Tommaso Fornaciari. 2020.
\newblock \href {https://doi.org/10.18653/v1/2020.acl-main.154} {{``}you sound
  just like your father{''} commercial machine translation systems include
  stylistic biases}.
\newblock In \emph{Proceedings of the 58th Annual Meeting of the Association
  for Computational Linguistics}, pages 1686--1690, Online. Association for
  Computational Linguistics.

\bibitem[{Hovy and Yang(2021)}]{hovy-yang-2021-importance}
Dirk Hovy and Diyi Yang. 2021.
\newblock \href {https://www.aclweb.org/anthology/2021.naacl-main.49} {The
  importance of modeling social factors of language: Theory and practice}.
\newblock In \emph{Proceedings of the 2021 Conference of the North American
  Chapter of the Association for Computational Linguistics: Human Language
  Technologies}, pages 588--602, Online. Association for Computational
  Linguistics.

\bibitem[{Jhamtani et~al.(2017)Jhamtani, Gangal, Hovy, and
  Nyberg}]{jhamtani-etal-2017-shakespearizing}
Harsh Jhamtani, Varun Gangal, Eduard Hovy, and Eric Nyberg. 2017.
\newblock \href {https://doi.org/10.18653/v1/W17-4902} {Shakespearizing modern
  language using copy-enriched sequence to sequence models}.
\newblock In \emph{Proceedings of the Workshop on Stylistic Variation}, pages
  10--19, Copenhagen, Denmark. Association for Computational Linguistics.

\bibitem[{John et~al.(2019)John, Mou, Bahuleyan, and
  Vechtomova}]{john-etal-2019-disentangled}
Vineet John, Lili Mou, Hareesh Bahuleyan, and Olga Vechtomova. 2019.
\newblock \href {https://doi.org/10.18653/v1/P19-1041} {Disentangled
  representation learning for non-parallel text style transfer}.
\newblock In \emph{Proceedings of the 57th Annual Meeting of the Association
  for Computational Linguistics}, pages 424--434, Florence, Italy. Association
  for Computational Linguistics.

\bibitem[{Kang et~al.(2019)Kang, Gangal, and Hovy}]{kang-etal-2019-male}
Dongyeop Kang, Varun Gangal, and Eduard Hovy. 2019.
\newblock \href {https://doi.org/10.18653/v1/D19-1179} {(male, bachelor) and
  (female, {P}h.{D}) have different connotations: Parallelly annotated
  stylistic language dataset with multiple personas}.
\newblock In \emph{Proceedings of the 2019 Conference on Empirical Methods in
  Natural Language Processing and the 9th International Joint Conference on
  Natural Language Processing (EMNLP-IJCNLP)}, pages 1696--1706, Hong Kong,
  China. Association for Computational Linguistics.

\bibitem[{Kang and Hovy(2021)}]{kang2019xslue}
Dongyeop Kang and Eduard Hovy. 2021.
\newblock \href {https://doi.org/10.18653/v1/2021.acl-long.185} {Style is {NOT}
  a single variable: Case studies for cross-stylistic language understanding}.
\newblock In \emph{Proceedings of the 59th Annual Meeting of the Association
  for Computational Linguistics and the 11th International Joint Conference on
  Natural Language Processing (Volume 1: Long Papers)}, pages 2376--2387,
  Online. Association for Computational Linguistics.

\bibitem[{Kazemi et~al.(2019)Kazemi, Iranmanesh, and
  Nasrabadi}]{kazemi2019style}
Hadi Kazemi, Seyed~Mehdi Iranmanesh, and Nasser Nasrabadi. 2019.
\newblock Style and content disentanglement in generative adversarial networks.
\newblock In \emph{2019 IEEE Winter Conference on Applications of Computer
  Vision (WACV)}, pages 848--856. IEEE.

\bibitem[{Kerinec et~al.(2018)Kerinec, Braud, and
  S{\o}gaard}]{kerinec-etal-2018-deep}
Emma Kerinec, Chlo{\'e} Braud, and Anders S{\o}gaard. 2018.
\newblock \href {https://doi.org/10.18653/v1/W18-5401} {When does deep
  multi-task learning work for loosely related document classification tasks?}
\newblock In \emph{Proceedings of the 2018 {EMNLP} Workshop {B}lackbox{NLP}:
  Analyzing and Interpreting Neural Networks for {NLP}}, pages 1--8, Brussels,
  Belgium. Association for Computational Linguistics.

\bibitem[{Koehn et~al.(2007)Koehn, Hoang, Birch, Callison-Burch, Federico,
  Bertoldi, Cowan, Shen, Moran, Zens, Dyer, Bojar, Constantin, and
  Herbst}]{koehn-etal-2007-moses}
Philipp Koehn, Hieu Hoang, Alexandra Birch, Chris Callison-Burch, Marcello
  Federico, Nicola Bertoldi, Brooke Cowan, Wade Shen, Christine Moran, Richard
  Zens, Chris Dyer, Ond{\v{r}}ej Bojar, Alexandra Constantin, and Evan Herbst.
  2007.
\newblock \href {https://www.aclweb.org/anthology/P07-2045} {{M}oses: Open
  source toolkit for statistical machine translation}.
\newblock In \emph{Proceedings of the 45th Annual Meeting of the Association
  for Computational Linguistics Companion Volume Proceedings of the Demo and
  Poster Sessions}, pages 177--180, Prague, Czech Republic. Association for
  Computational Linguistics.

\bibitem[{Li et~al.(2019)Li, Zhang, Gan, Cheng, Brockett, Dolan, and
  Sun}]{li-etal-2019-domain}
Dianqi Li, Yizhe Zhang, Zhe Gan, Yu~Cheng, Chris Brockett, Bill Dolan, and
  Ming-Ting Sun. 2019.
\newblock \href {https://doi.org/10.18653/v1/D19-1325} {Domain adaptive text
  style transfer}.
\newblock In \emph{Proceedings of the 2019 Conference on Empirical Methods in
  Natural Language Processing and the 9th International Joint Conference on
  Natural Language Processing (EMNLP-IJCNLP)}, pages 3304--3313, Hong Kong,
  China. Association for Computational Linguistics.

\bibitem[{Li et~al.(2018)Li, Jia, He, and Liang}]{li-etal-2018-delete}
Juncen Li, Robin Jia, He~He, and Percy Liang. 2018.
\newblock \href {https://doi.org/10.18653/v1/N18-1169} {Delete, retrieve,
  generate: a simple approach to sentiment and style transfer}.
\newblock In \emph{Proceedings of the 2018 Conference of the North {A}merican
  Chapter of the Association for Computational Linguistics: Human Language
  Technologies, Volume 1 (Long Papers)}, pages 1865--1874, New Orleans,
  Louisiana. Association for Computational Linguistics.

\bibitem[{Liu et~al.(2015)Liu, Gao, He, Deng, Duh, and
  Wang}]{liu-etal-2015-representation}
Xiaodong Liu, Jianfeng Gao, Xiaodong He, Li~Deng, Kevin Duh, and Ye-yi Wang.
  2015.
\newblock \href {https://doi.org/10.3115/v1/N15-1092} {Representation learning
  using multi-task deep neural networks for semantic classification and
  information retrieval}.
\newblock In \emph{Proceedings of the 2015 Conference of the North {A}merican
  Chapter of the Association for Computational Linguistics: Human Language
  Technologies}, pages 912--921, Denver, Colorado. Association for
  Computational Linguistics.

\bibitem[{Liu et~al.(2019)Liu, He, Chen, and Gao}]{liu-etal-2019-multi-task}
Xiaodong Liu, Pengcheng He, Weizhu Chen, and Jianfeng Gao. 2019.
\newblock \href {https://doi.org/10.18653/v1/P19-1441} {Multi-task deep neural
  networks for natural language understanding}.
\newblock In \emph{Proceedings of the 57th Annual Meeting of the Association
  for Computational Linguistics}, pages 4487--4496, Florence, Italy.
  Association for Computational Linguistics.

\bibitem[{Marcus et~al.(1993)Marcus, Marcinkiewicz, and
  Santorini}]{marcus1993building}
Mitchell~P. Marcus, Mary~Ann Marcinkiewicz, and Beatrice Santorini. 1993.
\newblock Building a large annotated corpus of english: The penn treebank.
\newblock \emph{Comput. Linguist.}, 19(2):313–330.

\bibitem[{McDonald et~al.(2013)McDonald, Nivre, Quirmbach-Brundage, Goldberg,
  Das, Ganchev, Hall, Petrov, Zhang, T{\"a}ckstr{\"o}m
  et~al.}]{mcdonald2013universal}
Ryan McDonald, Joakim Nivre, Yvonne Quirmbach-Brundage, Yoav Goldberg, Dipanjan
  Das, Kuzman Ganchev, Keith Hall, Slav Petrov, Hao Zhang, Oscar
  T{\"a}ckstr{\"o}m, et~al. 2013.
\newblock Universal dependency annotation for multilingual parsing.
\newblock In \emph{Proceedings of the 51st Annual Meeting of the Association
  for Computational Linguistics (Volume 2: Short Papers)}, pages 92--97.

\bibitem[{Mir et~al.(2019)Mir, Felbo, Obradovich, and
  Rahwan}]{mir-etal-2019-evaluating}
Remi Mir, Bjarke Felbo, Nick Obradovich, and Iyad Rahwan. 2019.
\newblock \href {https://doi.org/10.18653/v1/N19-1049} {Evaluating style
  transfer for text}.
\newblock In \emph{Proceedings of the 2019 Conference of the North {A}merican
  Chapter of the Association for Computational Linguistics: Human Language
  Technologies, Volume 1 (Long and Short Papers)}, pages 495--504, Minneapolis,
  Minnesota. Association for Computational Linguistics.

\bibitem[{Pan and Yang(2009)}]{pan2009survey}
Sinno~Jialin Pan and Qiang Yang. 2009.
\newblock A survey on transfer learning.
\newblock \emph{IEEE Transactions on knowledge and data engineering},
  22(10):1345--1359.

\bibitem[{Papineni et~al.(2002)Papineni, Roukos, Ward, and
  Zhu}]{papineni2002bleu}
Kishore Papineni, Salim Roukos, Todd Ward, and Wei-Jing Zhu. 2002.
\newblock {BLEU}: A method for automatic evaluation of machine translation.
\newblock In \emph{Proceedings of the 40th annual meeting of the Association
  for Computational Linguistics}, pages 311--318.

\bibitem[{Pavlick and Tetreault(2016)}]{pavlick-tetreault-2016-empirical}
Ellie Pavlick and Joel Tetreault. 2016.
\newblock \href {https://doi.org/10.1162/tacl_a_00083} {An empirical analysis
  of formality in online communication}.
\newblock \emph{Transactions of the Association for Computational Linguistics},
  4:61--74.

\bibitem[{Peterson et~al.(2011)Peterson, Hohensee, and
  Xia}]{peterson-etal-2011-email}
Kelly Peterson, Matt Hohensee, and Fei Xia. 2011.
\newblock \href {https://www.aclweb.org/anthology/W11-0711} {Email formality in
  the workplace: A case study on the {E}nron corpus}.
\newblock In \emph{Proceedings of the Workshop on Language in Social Media
  ({LSM} 2011)}, pages 86--95, Portland, Oregon. Association for Computational
  Linguistics.

\bibitem[{Poliak et~al.(2018)Poliak, Naradowsky, Haldar, Rudinger, and
  Van~Durme}]{poliak-etal-2018-hypothesis}
Adam Poliak, Jason Naradowsky, Aparajita Haldar, Rachel Rudinger, and Benjamin
  Van~Durme. 2018.
\newblock \href {https://doi.org/10.18653/v1/S18-2023} {Hypothesis only
  baselines in natural language inference}.
\newblock In \emph{Proceedings of the Seventh Joint Conference on Lexical and
  Computational Semantics}, pages 180--191, New Orleans, Louisiana. Association
  for Computational Linguistics.

\bibitem[{Prabhumoye et~al.(2018)Prabhumoye, Tsvetkov, Salakhutdinov, and
  Black}]{prabhumoye-etal-2018-style}
Shrimai Prabhumoye, Yulia Tsvetkov, Ruslan Salakhutdinov, and Alan~W Black.
  2018.
\newblock \href {https://doi.org/10.18653/v1/P18-1080} {Style transfer through
  back-translation}.
\newblock In \emph{Proceedings of the 56th Annual Meeting of the Association
  for Computational Linguistics (Volume 1: Long Papers)}, pages 866--876,
  Melbourne, Australia. Association for Computational Linguistics.

\bibitem[{Pryzant et~al.(2020)Pryzant, Martinez, Dass, Kurohashi, Jurafsky, and
  Yang}]{pryzant2020automatically}
Reid Pryzant, Richard~Diehl Martinez, Nathan Dass, Sadao Kurohashi, Dan
  Jurafsky, and Diyi Yang. 2020.
\newblock Automatically neutralizing subjective bias in text.
\newblock In \emph{Proceedings of the aaai conference on artificial
  intelligence}, volume~34, pages 480--489.

\bibitem[{Qi et~al.(2020)Qi, Zhang, Zhang, Bolton, and
  Manning}]{qi-etal-2020-stanza}
Peng Qi, Yuhao Zhang, Yuhui Zhang, Jason Bolton, and Christopher~D. Manning.
  2020.
\newblock \href {https://doi.org/10.18653/v1/2020.acl-demos.14} {{S}tanza: A
  python natural language processing toolkit for many human languages}.
\newblock In \emph{Proceedings of the 58th Annual Meeting of the Association
  for Computational Linguistics: System Demonstrations}, pages 101--108,
  Online. Association for Computational Linguistics.

\bibitem[{Radford et~al.(2019)Radford, Wu, Child, Luan, Amodei, Sutskever
  et~al.}]{radford2019language}
Alec Radford, Jeffrey Wu, Rewon Child, David Luan, Dario Amodei, Ilya
  Sutskever, et~al. 2019.
\newblock Language models are unsupervised multitask learners.
\newblock \emph{OpenAI blog}, 1(8):9.

\bibitem[{Raffel et~al.(2020)Raffel, Shazeer, Roberts, Lee, Narang, Matena,
  Zhou, Li, and Liu}]{JMLR:v21:20-074}
Colin Raffel, Noam Shazeer, Adam Roberts, Katherine Lee, Sharan Narang, Michael
  Matena, Yanqi Zhou, Wei Li, and Peter~J. Liu. 2020.
\newblock \href {http://jmlr.org/papers/v21/20-074.html} {Exploring the limits
  of transfer learning with a unified text-to-text transformer}.
\newblock \emph{Journal of Machine Learning Research}, 21(140):1--67.

\bibitem[{Rao and Tetreault(2018)}]{Tetreault2018}
Sudha Rao and Joel Tetreault. 2018.
\newblock \href {https://doi.org/10.18653/v1/N18-1012} {Dear sir or madam, may
  {I} introduce the {GYAFC} dataset: Corpus, benchmarks and metrics for
  formality style transfer}.
\newblock In \emph{Proceedings of the 2018 Conference of the North {A}merican
  Chapter of the Association for Computational Linguistics: Human Language
  Technologies, Volume 1 (Long Papers)}, pages 129--140, New Orleans,
  Louisiana. Association for Computational Linguistics.

\bibitem[{Roemmele et~al.(2017)Roemmele, Gordon, and
  Swanson}]{roemmele2017evaluating}
Melissa Roemmele, Andrew~S Gordon, and Reid Swanson. 2017.
\newblock Evaluating story generation systems using automated linguistic
  analyses.
\newblock In \emph{SIGKDD 2017 Workshop on Machine Learning for Creativity},
  pages 13--17.

\bibitem[{Ruder and Plank(2017)}]{ruder-plank-2017-learning}
Sebastian Ruder and Barbara Plank. 2017.
\newblock \href {https://doi.org/10.18653/v1/D17-1038} {Learning to select data
  for transfer learning with {B}ayesian optimization}.
\newblock In \emph{Proceedings of the 2017 Conference on Empirical Methods in
  Natural Language Processing}, pages 372--382, Copenhagen, Denmark.
  Association for Computational Linguistics.

\bibitem[{Nogueira~dos Santos et~al.(2018)Nogueira~dos Santos, Melnyk, and
  Padhi}]{nogueira-dos-santos-etal-2018-fighting}
Cicero Nogueira~dos Santos, Igor Melnyk, and Inkit Padhi. 2018.
\newblock \href {https://doi.org/10.18653/v1/P18-2031} {Fighting offensive
  language on social media with unsupervised text style transfer}.
\newblock In \emph{Proceedings of the 56th Annual Meeting of the Association
  for Computational Linguistics (Volume 2: Short Papers)}, pages 189--194,
  Melbourne, Australia. Association for Computational Linguistics.

\bibitem[{Schoch et~al.(2020)Schoch, Yang, and Ji}]{schoch-etal-2020-problem}
Stephanie Schoch, Diyi Yang, and Yangfeng Ji. 2020.
\newblock \href {https://www.aclweb.org/anthology/2020.evalnlgeval-1.2}
  {{``}{T}his is a problem, don{'}t you agree?{''} {F}raming and bias in human
  evaluation for natural language generation}.
\newblock In \emph{Proceedings of the 1st Workshop on Evaluating NLG
  Evaluation}, pages 10--16, Online (Dublin, Ireland). Association for
  Computational Linguistics.

\bibitem[{Shen et~al.(2017)Shen, Lei, Barzilay, and Jaakkola}]{shen2017style}
Tianxiao Shen, Tao Lei, Regina Barzilay, and Tommi Jaakkola. 2017.
\newblock Style transfer from non-parallel text by cross-alignment.
\newblock In \emph{Proceedings of the 31st International Conference on Neural
  Information Processing Systems}, pages 6833--6844.

\bibitem[{S{\o}gaard et~al.(2014)S{\o}gaard, Plank, and
  Hovy}]{sogaard-etal-2014-selection}
Anders S{\o}gaard, Barbara Plank, and Dirk Hovy. 2014.
\newblock \href {https://www.aclweb.org/anthology/C14-3005} {Selection bias,
  label bias, and bias in ground truth}.
\newblock In \emph{Proceedings of {COLING} 2014, the 25th International
  Conference on Computational Linguistics: Tutorial Abstracts}, pages 11--13,
  Dublin, Ireland. Dublin City University and Association for Computational
  Linguistics.

\bibitem[{Strunk and White(1999)}]{strunk_elements_1999}
William Strunk and E.~B. White. 1999.
\newblock \emph{The elements of style}, 4th ed edition.
\newblock Allyn and Bacon, Boston.

\bibitem[{Wang et~al.(2020)Wang, Che, Liu, Qin, Liu, and Wang}]{wang2020multi}
Shaolei Wang, Wangxiang Che, Qi~Liu, Pengda Qin, Ting Liu, and William~Yang
  Wang. 2020.
\newblock Multi-task self-supervised learning for disfluency detection.
\newblock In \emph{Proceedings of the AAAI Conference on Artificial
  Intelligence}, volume~34, pages 9193--9200.

\bibitem[{Wang et~al.(2021)Wang, Hoang, and Federico}]{wangtowards}
Yue Wang, Cuong Hoang, and Marcello Federico. 2021.
\newblock Towards modeling the style of translators in neural machine
  translation.
\newblock In \emph{Proceedings of the 2021 Conference of the North American
  Chapter of the Association for Computational Linguistics: Human Language
  Technologies}, pages 1193--1199.

\bibitem[{Wang et~al.(2019)Wang, Wu, Mou, Li, and
  Chao}]{wang-etal-2019-harnessing}
Yunli Wang, Yu~Wu, Lili Mou, Zhoujun Li, and Wenhan Chao. 2019.
\newblock \href {https://doi.org/10.18653/v1/D19-1365} {Harnessing pre-trained
  neural networks with rules for formality style transfer}.
\newblock In \emph{Proceedings of the 2019 Conference on Empirical Methods in
  Natural Language Processing and the 9th International Joint Conference on
  Natural Language Processing (EMNLP-IJCNLP)}, pages 3573--3578, Hong Kong,
  China. Association for Computational Linguistics.

\bibitem[{Williams and Bizup(2017)}]{williams_style_2017}
Joseph~M. Williams and Joseph Bizup. 2017.
\newblock \emph{Style: lessons in clarity and grace}, twelfth edition edition.
\newblock Pearson, Boston.

\bibitem[{Xu(2017)}]{xu-2017-shakespeare}
Wei Xu. 2017.
\newblock \href {https://doi.org/10.18653/v1/W17-4901} {From shakespeare to
  {T}witter: What are language styles all about?}
\newblock In \emph{Proceedings of the Workshop on Stylistic Variation}, pages
  1--9, Copenhagen, Denmark. Association for Computational Linguistics.

\bibitem[{Xu et~al.(2012)Xu, Ritter, Dolan, Grishman, and
  Cherry}]{xu-etal-2012-paraphrasing}
Wei Xu, Alan Ritter, Bill Dolan, Ralph Grishman, and Colin Cherry. 2012.
\newblock \href {https://www.aclweb.org/anthology/C12-1177} {Paraphrasing for
  style}.
\newblock In \emph{Proceedings of {COLING} 2012}, pages 2899--2914, Mumbai,
  India. The COLING 2012 Organizing Committee.

\bibitem[{Zhang et~al.(2020)Zhang, Ge, and Sun}]{zhang-etal-2020-parallel}
Yi~Zhang, Tao Ge, and Xu~Sun. 2020.
\newblock \href {https://doi.org/10.18653/v1/2020.acl-main.294} {Parallel data
  augmentation for formality style transfer}.
\newblock In \emph{Proceedings of the 58th Annual Meeting of the Association
  for Computational Linguistics}, pages 3221--3228, Online. Association for
  Computational Linguistics.

\end{thebibliography}

\clearpage
\appendix

\section{Dataset Details}\label{sec:datasets}
We selected English text style datasets with a single transferred stylistic attribute between two classes. Of importance for inclusions were datasets that exhibited different creation methods: both automatically annotated and human annotated. Where available, we used the original (or pre-existing, as with the case of the Shakespeare dataset) train/val/test data splits. Links to each dataset are provided through the respective citations.

\paragraph{Fluency} Contains aligned sentence pairs labeled as fluent or disfluent, from the English Switchboard (SWBD) Corpus \citep{godfrey1992switchboard, wang2020multi}. Train/val/test split: 173.7k/10.1k/7.9k

\paragraph{GYAFC-EM \& GYAFC-FR} Contain aligned sentence pairs labeled as informal or formal, from the \emph{Entertainment \& Music} and \emph{Family \& Relationships} domains, respectively, of the question answering forum Yahoo Answers \citep{Tetreault2018}. GYAFC-EM \& GYAFC-FR datasets can be requested at \url{https://github.com/raosudha89/GYAFC-corpus}. GYAFC-EM Train/val/test split: 52.6k/2.9k/1.4k; GYAFC-FR Train/val/test split: 52k/2.8k/1.3k

\paragraph{Biased-Word} Contains aligned sentence pairs labeled as subjective or neutral, crawled from 423,823 Wikipedia editor neutrilization revisions between 2004 and 2019 \citep{pryzant2020automatically}. Train/val/test split: 53.8k/700/1k

\paragraph{Flickr} Contains sentence pairs captioning an image, labeled as romantic or humorous \citep{gan2017stylenet}. We created a 6k/500/500 Train/val/test split since only the original 7k training instances are available.

\paragraph{Shakespeare} Contains sentence pairs labeled as Shakespeare or modern English \citep{xu-etal-2012-paraphrasing}. Sentences are crawled from 17 Shakespeare plays from Sparknotes \footnote{\url{https://www.sparknotes.com/}}, which provides the modern counterparts. Following \citet{jhamtani-etal-2017-shakespearizing}, we use 15 plays for training, with \emph{Twelfth Night} used for validation, and \emph{Romeo and Juliet} used for testing.

\section{Similarity Metrics}\label{sec:similarity-app}
In Table \ref{sent-sims-table} we do not distinguish between source and target direction due to the symmetry of metrics in our setting. We provide further justification below:

Jaccard similarity can be defined as

\begin{equation}
\label{eq:jacsim}
    \frac{\mathcal{V}_{\{\vec{s}^{(k)}\}} \cap \mathcal{V}_{\{\vec{t}^{(k)}\}}}{\mathcal{V}_{\{\vec{s}^{(k)}\}} \cup \mathcal{V}_{\{\vec{t}^{(k)}\}}}
\end{equation}

where $\mathcal{V}_{\{\vec{s}^{(k)}\}}$ denotes the set of vocabulary words existing in a source sentence $\{\vec{s}^{(k)}\}$ and  $\mathcal{V}_{\{\vec{t}^{(k)}\}}$ denotes the set of vocabulary words existing in a target sentence $\{\vec{t}^{(k)}\}$. By the commutative property, $\mathcal{V}_{\{\vec{s}^{(k)}\}} \cap \mathcal{V}_{\{\vec{t}^{(k)}\}} = \mathcal{V}_{\{\vec{t}^{(k)}\}} \cap \mathcal{V}_{\{\vec{s}^{(k)}\}}$ and $\mathcal{V}_{\{\vec{s}^{(k)}\}} \cup \mathcal{V}_{\{\vec{t}^{(k)}\}} = \mathcal{V}_{\{\vec{t}^{(k)}\}} \cup \mathcal{V}_{\{\vec{s}^{(k)}\}}$, making Jaccard similarity symmetric. Word-based Levenshtein distance is defined as the minimum number of edit operations to convert $\{\vec{s}^{(k)}\}$ to $\{\vec{t}^{(k)}\}$ through insertions, deletions, and substitutions. Substitutions are symmetric by definition, and insert and delete operations to convert $\{\vec{s}^{(k)}\}$ to $\{\vec{t}^{(k)}\}$ are simply reversed when converting $\{\vec{t}^{(k)}\}$ to $\{\vec{s}^{(k)}\}$. In $LD_{norm}(s,t)$, we normalize by $\max|s|,|t|$, which is invariant to order. Finally,

\begin{equation}
\label{eq:f1}
    F_1 = 2 * \frac{\textit{precision}*\textit{recall}}{\textit{precision}+\textit{recall}}
\end{equation}

where $\textit{precision} = \frac{\textit{TP}}{\textit{TP}+\textit{FP}}$ and $\textit{recall} = \frac{\textit{TP}}{\textit{TP}+\textit{FN}}$.

In our setting, $\textit{TP}=w\in s \cap t$ , $\textit{FP}=w \in \vec{s}\backslash\vec{t}$, and $\textit{FN} = w \in \vec{t}\backslash\vec{s}$. By these definitions, $\textit{FP}$ and $\textit{FN}$ are reversed when source and target are reversed, and therefore by definition, $F_1$ is symmetric when comparing source and target sentence pairs.\footnote{Acronyms refer to ``True Positives'' (TP), ``False Positives'' (FP), and ``False Negatives'' (FN). We consider target as ground truth and copy source over as a ``generated'' target. We essentially consider positives as words that are generated and negatives as words that are not generated, with truth values corresponding to whether or not a word \textit{should} have been generated.}

\section{Linguistic Features}\label{ling-feat}
\paragraph{Lexical Complexity} Lexical complexity refers to the complexity of words based on the length or number of syllables. We use average word length in characters \citep{pavlick-tetreault-2016-empirical} and average number of syllables, with and without stopwords.

\paragraph{Lexical Diversity} Size of vocabulary has been used as a feature for style categorization in prior work \citep{abu-jbara-etal-2011-towards}. We chose to include unigrams and bigrams to reflect diversity of vocabulary as well as diversity of expression.

\paragraph{POS Tags} POS tags have been used extensively in the stylistic analysis of text, including formality \citep{pavlick-tetreault-2016-empirical} and written-style vs. audio-style \citep{abu-jbara-etal-2011-towards}. Granularity of POS tags has stylistic implications, such as implications for different specific punctuation types \citep{strunk_elements_1999}, so we include Universal and Treebank POS tags for course-grained and fine-grained stylistic information, respectively. \footnote{Although we used state-of-the-art tools to extract features such as part-of-speech tags, we do note the possibility of tool performance differences across datasets \citep{sogaard-etal-2014-selection}. However, as we utilize the same tool for both the classification and ablation study as well as the divergence scores, we expect the impact of tool performance within a dataset to have minimal impact on resulting conclusions.}

Both Universal and Treebank POS tags are processed using Stanza \citep{qi-etal-2020-stanza}, which correspond with the Universal Dependencies \citep{mcdonald2013universal} POS tags and the Penn Treebank \citep{marcus1993building} English POS tagset.

\paragraph{Sentence Length} Sentence length has stylistic implications \citep{strunk_elements_1999} and has been used as a feature to classify various styles, such as written-style and audio style \citep{abu-jbara-etal-2011-towards} and formality \citep{pavlick-tetreault-2016-empirical}. We include sentence length in words and sentence length in tokens to account for punctuation differences. 

\paragraph{Phrases} Measures of phrases and clauses have been used for stylistic analysis in terms of syntactic complexity \citep{abu-jbara-etal-2011-towards}. We include measures of noun phrases, verb phrases, and dependent clauses.

\paragraph{Readability} We adopt the readability measures Flesch-Kincaid Grade Level score \citep{pavlick-tetreault-2016-empirical} and ratio of complex words \citep{abu-jbara-etal-2011-towards} from prior studies.

\paragraph{Subjectivity} We adopted several measures of subjectivity from \citet{pavlick-tetreault-2016-empirical}
and adapted the measure ratio of pronouns \citep{abu-jbara-etal-2011-towards} by measuring the individual type counts of 1st, 2nd, and 3rd person pronouns.

\paragraph{Bag-of-Words} We include the bag-of-words feature to account for cross-class vocabulary differences.

\section{Jensen-Shannon Divergence}\label{div-app}
While we indicate large Jensen-Shannon Divergences in Table \ref{heatmap}, we include the full range of Jensen-Shannon Divergence results in Table \ref{divergence-table} in a numerical format as well.

\begin{table}[h]
\resizebox{\columnwidth}{!}{%
    \small
    \centering
    \begin{tabular}{r*{8}{r}}
        \multicolumn{1}{c}{} &
        \multicolumn{1}{c}{Flick} &
        \multicolumn{1}{c}{Shake} &
        \multicolumn{1}{c}{GY-FR} &
        \multicolumn{1}{c}{GY-EM} & \multicolumn{1}{c}{Bias} & \multicolumn{1}{c}{Flu.}\\ 
        FF&\gradientc{0.022}&\gradientc{0.019}&\gradientc{0.019}&\gradientc{0.027}&\gradientc{0.003}&\gradientc{0.004}\\
        LexC&\textbf{\gradientc{0.086}}&\gradientc{0.039}&\textbf{\gradientc{0.132}}&\gradientc{0.054}&\gradientc{0.047}&\gradientc{0.004}\\
        Read&\textbf{\gradientc{0.081}}&\gradientc{0.040}&\textbf{\gradientc{0.079}}&\gradientc{0.056}&\gradientc{0.050}&\gradientc{0.013}\\ 
        LexD&\gradientc{0.067}&\gradientc{0.049}&\gradientc{0.041}&\gradientc{0.050}&\gradientc{0.031}&\textbf{\gradientc{0.108}}\\ 
        UPOS&\textbf{\gradientc{0.088}}&\gradientc{0.052}&\gradientc{0.066}&\textbf{\gradientc{0.075}}&\gradientc{0.034}&\gradientc{0.011}\\ 
        XPOS&\gradientc{0.063}&\gradientc{0.042}&\gradientc{0.052}&\gradientc{0.056}&\gradientc{0.026}&\gradientc{0.008}\\
        SenL&\textbf{\gradientc{0.137}}&\textbf{\gradientc{0.090}}&\gradientc{0.070}&\gradientc{0.062}&\gradientc{0.013}&\gradientc{0.017}\\
        Phr&\textbf{\gradientc{0.105}}&\gradientc{0.056}&\gradientc{0.064}&\gradientc{0.065}&\gradientc{0.030}&\gradientc{0.024}\\
        Sub&\textbf{\gradientc{0.107}}&\textbf{\gradientc{0.075}}&\gradientc{0.054}&\gradientc{0.057}&\gradientc{0.064}&\gradientc{0.016}\\
        BoW&\gradientc{0.018}&\gradientc{0.015}&\gradientc{0.013}&\gradientc{0.011}&\gradientc{0.002}&\gradientc{0.002}\\ 
    \end{tabular}}

\caption{\label{divergence-table}Jensen-Shannon divergence between source and target on each test set using feature groupings in Table \ref{features-table}. Scores $\geq 0.075$ are made bold.}
\end{table}

\end{document}